\documentclass[letterpaper,12pt]{article}

\usepackage[english]{babel}
\usepackage[utf8]{inputenc}
\usepackage[normalem]{ulem}

\usepackage{amsmath,amssymb,amsfonts,amsthm,mathtools,bm}
\usepackage{cases,empheq}

\usepackage{graphicx}
\usepackage{booktabs,multirow,array}
\usepackage[font=small,labelfont=bf]{caption}
\usepackage{subcaption} 

\usepackage[ruled,linesnumbered,vlined]{algorithm2e}

\usepackage{xcolor}
\usepackage{tcolorbox}
\usepackage{verbatim}
\usepackage{setspace}
\usepackage{indentfirst}
\usepackage{titlesec,titletoc}
\usepackage{fancyhdr}
\usepackage{wasysym}
\usepackage{natbib}
\usepackage{chngcntr}

\usepackage{enumitem}

\usepackage{tikz}

\usepackage{hyperref} 
\usepackage{xr}            
\newtheorem{theorem}[]{Theorem}

\newtheorem{proposition}[]{Proposition}

\newtheorem{assumption}[]{Assumption}

\newcommand{\argmax}{\operatornamewithlimits{argmax}}

\def\({\left(}\def\){\right)}\def\[{\left[}\def\]{\right]}

\newcommand{\rom}[1]{\uppercase\expandafter{\romannumeral #1\relax}}

\newcommand{\blind}{0} 

\begin{document}

\def\spacingset#1{\renewcommand{\baselinestretch}%
{#1}\small\normalsize} \spacingset{1}

\if0\blind
{
  \title{\bf Learn then Decide: A Learning Approach for Designing Data Marketplaces}
  \author{Yingqi Gao\textsuperscript{1}, Wenlu Xu\textsuperscript{1},  Jin J. Zhou\textsuperscript{2}, \\ Hua Zhou\textsuperscript{2}, Yong Chen\textsuperscript{3}, and Xiaowu Dai\textsuperscript{1,2,$*$}\\
  \textsuperscript{1} \it \normalsize Department of Statistics and Data Science, University of California, Los Angeles\\
  \textsuperscript{2} \it \normalsize Department of Biostatistics, University of California, Los Angeles\\
  \textsuperscript{3} \it \normalsize Department of Biostatistics, University of Pennsylvania}
  \date{}
  \maketitle
} \fi

\if1\blind
{
  \bigskip
  \bigskip
  \bigskip
  \begin{center}
    {\LARGE Learn then Decide: A Learning Approach for Designing Data Marketplaces}
\end{center}
  \medskip
} \fi

\if0\blind
{
\begin{footnotetext}[1]
{\textit{Address for correspondence:} Xiaowu Dai, Department of Statistics and Data Science and Department of Biostatistics, UCLA, 8917 Math Sciences Bldg \#951554,  Los Angeles, CA 90095,  USA. Email: dai@stat.ucla.edu.}
\end{footnotetext}
} \fi

\smallskip

\begin{center}
  To appear in   \emph{Journal of the American Statistical Association}.
\end{center}
  
\bigskip
\begin{abstract}
\noindent
As data marketplaces become increasingly central to the digital economy, it is crucial to design efficient pricing mechanisms that optimize revenue while ensuring fair and adaptive pricing. 
We introduce the Maximum Auction-to-Posted Price (MAPP) mechanism,
a novel two-stage approach that first estimates the bidders’ value distribution through
auctions and then determines the optimal posted price based on the learned distribution. 
We establish that MAPP  is individually rational and incentive-compatible,
ensuring truthful bidding while balancing revenue maximization with minimal price
discrimination. 
On the theoretical side, we establish a statistical viewpoint that 
recasts revenue optimization as a valuation density estimation problem: 
we show that revenue regret can be controlled by uniform error in estimating the valuation density.
MAPP achieves a regret of $O_p(n^{-1}(\log n)^2)$ when incorporating historical bid
data, where $n$  is the number of bids in the current round.
For sequential dataset sales over $T$ rounds, we propose an online MAPP mechanism that dynamically adjusts pricing across datasets with varying value distributions. Our approach achieves no-regret
learning, with the average cumulative regret converging at a rate of
 $O_p(T^{-1/2}(\log T)^2)$.
 We validate the effectiveness of MAPP through simulations and real-world data from
the FCC AWS\textendash 3 spectrum auction.
\end{abstract}
\bigskip

\noindent%
\vfill

\newpage
\spacingset{1.9} 
\section{Introduction}
\label{sec:intro}
\noindent
The economic value of data has grown rapidly in recent decades,
making it a key resource for artificial intelligence and decision making \citep{VeldkampChung2024}. 
The increasing demand for data has led to the emergence of data marketplaces 
such as Amazon AWS Data Exchange, Datarade, and Snowflake Data Marketplace. 
However, these platforms remain inefficient, because 
transactions often involve prolonged price negotiations that generate high transaction costs \citep{KennedyEtAl2022}. This inefficiency motivates a natural question: Can
alternative pricing mechanisms be designed to not only maximize seller revenue but also
enhance the efficiency of data trading?

A primary driver of prolonged negotiation is asymmetric information about buyer demand. Auctions provide a natural tool to address this, since well-designed auctions both allocate efficiently and reveal how much buyers are willing to pay, which the seller can use to optimize revenue \citep{Myerson1981, Segal2003, KanoriaNazerzadeh2021} to optimize revenue. Relying exclusively on auctions, however, is poorly suited to data marketplaces with a long stream of arriving buyers, because a one-shot auction serves only participants in that round. Recent empirical evidence shows that auctions are losing favor in real-world platforms such as eBay, where sellers increasingly favor posted prices in response to buyers’ preference for convenience; yet posted prices reveal little information about demand to the seller \citep{EinavEtAl2018}. This shift highlights the need for hybrid mechanisms that combine the information-revealing and revenue-efficient properties of auctions with the simplicity and accessibility of posted pricing.

\begin{figure}[ht!]
  \centering
  \includegraphics[width=0.85\textwidth]{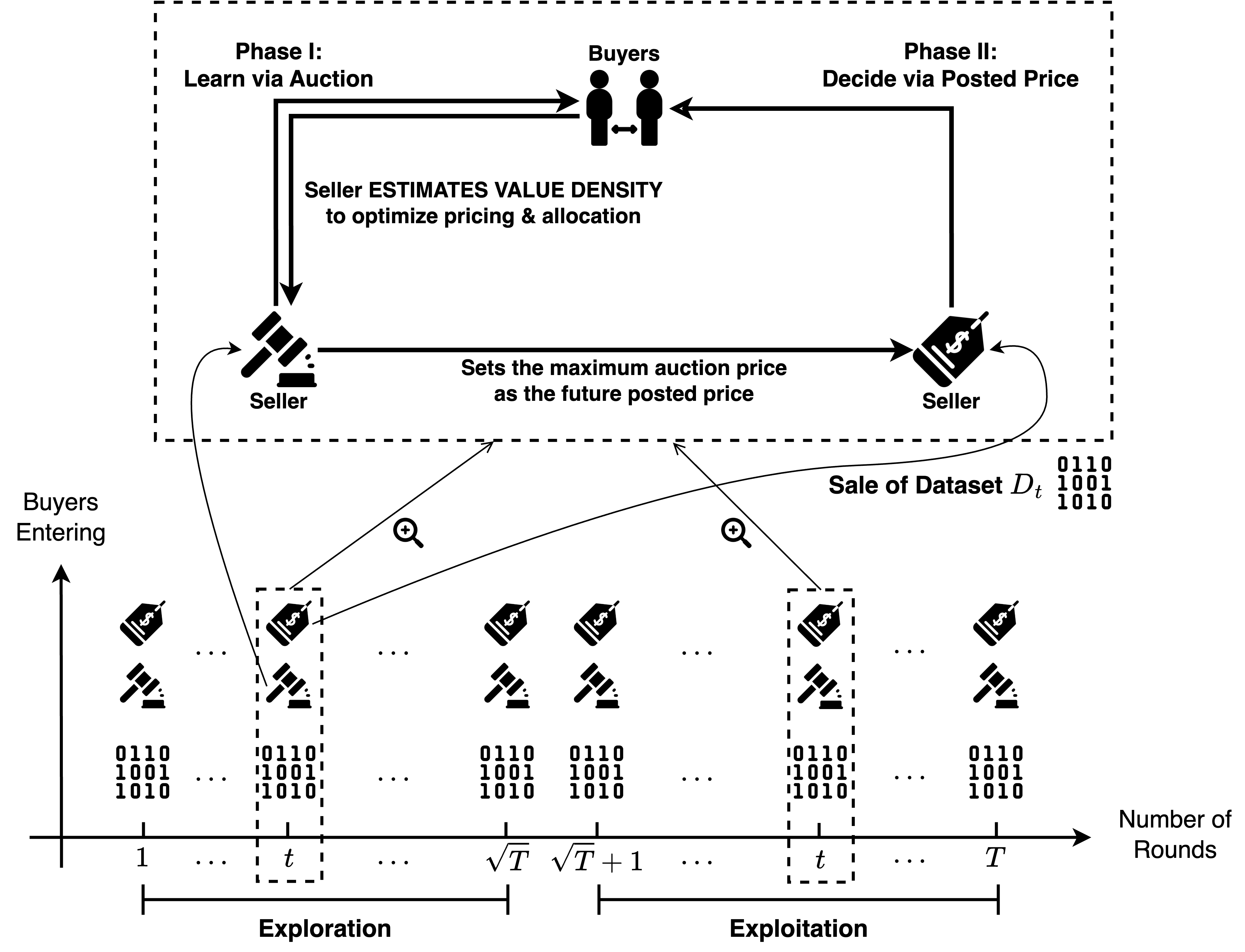}
  \caption{llustration of the online MAPP mechanism operating over T rounds.}
  \label{fig:MAPP}
\end{figure}

This paper introduces the Maximum Auction-to-Posted Price (MAPP) mechanism, a two-phase pricing strategy that begins with an auction phase and then uses a posted-price phase in order to learn demand and maximize revenue in data trading. The workflow of MAPP is illustrated in Figure~\ref{fig:MAPP}. In the auction phase, the seller runs a bid-independent auction among the buyers who are currently present: buyers submit bids and are then offered prices that depend only on the bids of others. Any buyer whose bid is at least her assigned price is required to purchase the dataset. This bid-independent structure ensures incentive compatibility, meaning that truthful bidding is each buyer’s optimal strategy. Moreover, it guarantees that no buyer pays more than her bid and reduces the degree of price discrimination by grouping buyers into only two pricing cohorts. In the subsequent posted-price phase, the seller posts the maximum auction price as a take-it-or-leave-it offer to future buyers. Buyers accept the offer only if their valuations exceed the posted price, ensuring individual rationality; that is, participation is voluntary, which is essential for market efficiency.

A key factor governing the performance of MAPP is how accurately it can estimate the valuation distribution from observed bids in order to set a near-optimal price. If the distribution were known, the optimal price would follow classical auction theory \citep{Myerson1981}. In practice, however, uncertainty about the valuation distribution leads to  regret, which measures the resulting loss in revenue. We establish a general regret-density connection showing that revenue regret is controlled by the uniform estimation error of the valuation density, and thus reframing revenue optimization as a nonparametric density estimation problem. Although prior work has employed nonparametric estimators such as the empirical cumulative distribution function (eCDF) and kernel density estimation (KDE) \citep{Segal2003, GoldbergEtAl2006, ColeRoughgarden2014}, the connection between density estimation error and revenue regret is unclear. Our result makes this connection explicit and turns MAPP into a modular statistical framework in which any suitable density estimator can be incorporated and analyzed through the developed regret-density connection. Aligned with \citet{Segal2003}, we begin with the KDE approach. By combining the uniform KDE bounds of \citet{Jiang2017} with our regret–density framework, we obtain a finite-sample regret bound for MAPP that achieves the classical rate while requiring weaker regularity assumptions than existing auction analyses. MAPP also mitigates the price discrimination inherent in the price mechanism of \citet{Segal2003}.

When datasets arise from a common source and historical bids are available, we further improve the regret bound  using repeated density estimation (RDE) \citep{QiuEtAl2022}, which pools information across multiple selling rounds. By improving the existing $L_1$ error bound for RDE to uniform $L_\infty$ bound and combining it with our regret-density connection, we obtain a regret rate of order $O_p(n^{-1}(\log n)^2)$, a significant improvement over the existing rate $O(n^{-2/3}(\log n)^{3/2})$ \citep{Segal2003, KleinbergLeighton2003}, where $n$ is the number of bids for the current dataset.
Moreover, we extend MAPP to an online setting involving $T$ rounds of sequential datasets and  obtain a regret  $O_p(T^{-1/2} (\log T)^2)$. This establishes no-regret learning, in the sense that the average regret converges to zero as the number of rounds increases. The online MAPP mechanism consists of two stages.  In the exploration stage, the mechanism applies KDE-based estimation to the first $\sqrt{T}$ datasets. In the exploitation stage, pricing for the remaining datasets is refined using RDE, which pools information from the exploration stage  to produce more accurate valuation distribution estimates.
Simulation and real application using FCC AWS–3 auction data show that MAPP achieves the lowest regret and the smallest variance compared to alternative mechanisms across various market environments and bidder conditions.

Although we present MAPP in the context of data marketplaces, the same non-rivalrous, unlimited-supply structure also appears in other digital goods settings, and we discuss these broader applications in Appendix. The problem of designing revenue-optimal mechanisms for unlimited-supply markets, such as data products that can be sold to any number of buyers, has been studied. In particular, \citet{balcan2008reducing} introduce a learning-based framework that reduces incentive-compatible mechanism design to an algorithmic learning problem. Their approach shows how learning methods can be used in mechanism design without imposing strong distributional assumptions, in single-round auctions with sample-complexity guarantees.
We complement this line of work by building on this insight in a different regime: a dynamic data marketplace in which the seller learns across successive sales of heterogeneous datasets and performance is evaluated via no-regret guarantees. Rather than providing a general reduction, MAPP specifies a concrete two–phase mechanism—an initial auction followed by posted pricing—that learns buyer valuations dynamically while preserving incentive compatibility and individual rationality. This design enables adaptive revenue maximization, and our analysis establishes the first regret guarantee for such a hybrid auction–posted–price mechanism.

The rest of this paper is organized as follows. 
Section~\ref{sec:problem} introduces the data marketplace model and formalizes the incentive and regret criteria. 
Section~\ref{sec:mapp} presents the MAPP mechanism, establishes its incentive guarantees, and 
extends it to an online setting.  
Section~\ref{sec:theoretical} develops regret bounds for both instantaneous and cumulative regret. 
Section~\ref{sec:empirical} reports numerical experiments with simulated and real data. 
Section~\ref{sec:related} reviews related literature.
Section~\ref{sec:conclusion} concludes with discussions.
\section{Background}\label{sec:problem}

\subsection{Model Setup}
\label{sec:setup}
\noindent
We consider a data marketplace where a seller continuously offers datasets generated from a common data source, 
denoted by $\mathcal{D}$, and released sequentially over time. 
At each round $t = 1, \dots, T$, a new dataset $D^{(t)}$ from the source $\mathcal{D}$ becomes available for sale. 
Upon its release, $n^{(t)}$ potential buyers are present in the market, 
each deciding whether to make a purchase within a limited time window before exiting the market. 

To address ex-ante unverifiability and free duplication \citet{ChenEtAl2022}, 
the seller provides only a coarse signal of the dataset $D^{(t)}$, denoted as $\tilde{D}^{(t)}$,
such as data summaries or a small subset, instead of a full access \citep{ChenEtAl2022, GuoEtAl2022}. 
Given $\tilde{D}^{(t)}$, each buyer $i$ forms a posterior belief about the value of $D^{(t)}$ and 
derives a valuation
$v_i^{(t)} = \mathbb{E}[V_i(D^{(t)})| \tilde{D}^{(t)}],$
where $V_i(D^{(t)})$ denotes buyer $i$’s utility from accessing the complete dataset $D^{(t)}$ and 
the expectation is taken with respect to the buyer’s posterior distribution over $D^{(t)}$ conditional on $\tilde{D}^{(t)}$.
The valuations $v_i^{(t)}$, for $i=1,\ldots,n^{(t)}$, differ across buyers because of heterogeneous utilities. We model this variability by assuming that $v_i^{(t)}$s in round $t$ are independent draws from an unknown density $f^{(t)}$. Although $f^{(t)}$ may differ across round $t=1,\ldots,T$, each density belongs to a common family $\mathcal{F}$ as follows.

\begin{assumption}[Shared Density Family]
\label{assump:family}
For each round $t = 1,\dots,T$, the buyer valuations $v_i^{(t)}$ for dataset $D^{(t)}$, with $i=1,\ldots,n^{(t)}$, are i.i.d.\ draws from a Lipschitz continuous density $f^{(t)}$ supported on $[1,H]$, where $H < \infty$ is known to the seller and $f^{(t)}(v)>0$ for all $v \in [1,H]$. The densities $\{f^{(t)}\}_{t=1}^T$ consists of i.i.d.\ realizations from a common family $\mathcal{F}$.
\end{assumption}

\noindent
Assumption \ref{assump:family} aligns with real-world scenarios, such as Quora generates datasets $D^{(t)}$ formed from user-submitted questions and answers that reflect relatively stable topical patterns, due to content-moderation policies  \citep{wang2013wisdom}. As a result, variation across successive releases can be captured by gradual shifts in distributional parameters rather than structural changes in the data-generating process, which supports modeling $f^{(t)}$ as belonging to a common family $\mathcal{F}$.  The bounded support assumption $[1,H]$ is widely used in auctions \citep[e.g.,][]{GoldbergEtAl2006, HuangEtAl2015, GuoEtAl2022}. The Lipschitz continuity  imposes mild smoothness condition, ensuring that small changes in dataset quality translate into gradual shifts in valuations \citep{GuerreEtAl2000}.

\subsection{Incentive Measures}
\label{sec:incmeasure}
\noindent
This paper designs a mechanism for data marketplaces that enables the seller to effectively price and sell datasets at each round~$t$ to arriving buyers. Since the mechanism operates identically in each round, we fix any $t\in\{1,\ldots,T\}$ and omit the round index in this subsection. The mechanism integrates two components: an auction phase, which elicits bids and captures market information, and a posted-price phase, which offers a uniform price to subsequent buyers. This hybrid structure reflects the model setup in Section~\ref{sec:setup}, where buyers face limited waiting periods and must decide whether to purchase before exiting the market. It also mirrors a broader real-world trend in which markets have shifted from auctions to posted pricing, driven by increasing buyer impatience and the seller’s preference for operational simplicity \citep{EinavEtAl2018}.

For a mechanism to function effectively, it must satisfy key incentive properties that guarantee buyer participation and truthful bidding. In the MAPP framework, 
these properties are individual rationality (IR) and incentive compatibility (IC). 
While the concepts originate from classical mechanism design \citep{Vickrey1961,Groves1973,Myerson1981}, 
we adapt their definitions to our two-phase setting of an auction followed by posted pricing.

\noindent\textbf{Pricing functions.}  
Let $\mathbf{b}=(b_1,\dots,b_n)$ denote the submitted bids.  
The auction pricing rule $h_A$ maps $\mathbf{b}$ to the price vector $\mathbf{p}=(p_1,\dots,p_n)$, 
where $p_i = h_A(\mathbf{b})_i$ denotes the pricing to buyer $i$.  
Equivalently, $p_i = h_A(b_i,\mathbf{b}_{-i})_i$, 
where $\mathbf{b}_{-i}$ is the vector of bids from all buyers except~$i$.  
Later we focus on mechanisms where $h_A$ satisfies the standard bid-independence property  
so that $p_i$ depends only on $\mathbf{b}_{-i}$ \citep{GoldbergEtAl2006}.
After the auction phase, the mechanism computes a uniform posted price through 
a pricing function $h_P$, which maps the auction price vector $\mathbf{p}$ to the posted price $p=h_P(\mathbf{p})$.

\noindent\textbf{Individual Rationality (IR).}  
IR requires that participation does not yield negative utility. 
In the MAPP mechanism, the utility of buyer~$i$ in the auction phase is
$u_i(b_i, p_i) = (v_i - p_i)\cdot \mathbf{1}\{b_i \geq p_i\}$,
where $v_i$ is the buyer’s valuation. 
If buyer $i$ fails to buy in the auction, she may buy at posted price $p$,
with utility $u_i(v_i, p) = (v_i - p)\,\mathbf{1}\{v_i \geq p\}$.
The mechanism is IR if
$u_i(v_i, p_i) \;\geq\; u_i(v_i, p) \;\geq\; 0$,
so buyers never incur a loss from participation and have no incentive to delay purchase.

\noindent\textbf{Incentive Compatibility (IC).}  
IC requires that truthful bidding maximizes each buyer's expected utility.
For fixed $\mathbf{b}_{-i}$, buyer $i$'s expected utility from bidding $b_i$ is
$\bar{u}_i(b_i, \mathbf{b}_{-i}) 
    = u_i\bigl(b_i, h_A(b_i, \mathbf{b}_{-i})_i\bigr)
    + \mathbf{1}\{b_i < h_A(b_i, \mathbf{b}_{-i})_i\}\,
      u_i\bigl(v_i, h_P(h_A(b_i, \mathbf{b}_{-i}))\bigr)$.
The mechanism is IC if
$\bar{u}_i(v_i, \mathbf{v}_{-i}) 
    \;\ge\; \bar{u}_i(b_i, \mathbf{v}_{-i})~ 
    \forall\, b_i \in \mathbb{R}_+$,
so truthful bidding weakly dominates all deviations.
This extends the standard IC condition \citep{Myerson1981} 
to our two-phase mechanism.

\subsection{Regret Measure}
\label{sec:regretmeasure}
\noindent
We evaluate the performance of a mechanism in data marketplaces using regret, which
measures the gap between the revenue achieved by the mechanism and the optimal revenue
that would be obtained with perfect knowledge of the buyers’ value distribution.

\noindent\textbf{Instantaneous regret.} 

For dataset $D^{(t)}$, the instantaneous regret at price $p^{(t)}$ is
\begin{equation}
\label{eqn:definstreg}  
    r^{(t)}(p^{(t)}) = {OPT}^{(t)} - {REV}^{(t)}(p^{(t)}).
\end{equation}
Valuations follow a density $f^{(t)}$ on $[1,H]$ with CDF $F^{(t)}(p)=\int_{1}^{p} f^{(t)}(v)\,dv$. 
For any price $p$, the expected revenue is $REV^{(t)}(p)=p\{1-F^{(t)}(p)\}$, and 
$OPT^{(t)}=\max_{p\in[1,H]} REV^{(t)}(p)$ is the maximal expected revenue under the true valuation distribution.

\noindent\textbf{Convex-hull (ironed) revenue.}
Although $OPT^{(t)}$ is defined in price space, 
the maximization can be difficult when $f^{(t)}$ is multimodal and $REV^{(t)}(p)$ has multiple local maxima. 
To obtain a benchmark that is globally optimizable, it is convenient to reparameterize the problem in quantile space. Let $q = 1 - F^{(t)}(p)$ denote the sale probability associated with price and define the corresponding revenue curve
$REV_q^{(t)}(q) = q\,F^{(t),-1}(1-q)$, for $q \in [0,1]$,
where $F^{(t),-1}$ is the generalized inverse CDF given by $F^{(t),-1}(u) = \inf\{p \in [1,H] : F^{(t)}(p) \ge u\}$.
Under Assumption~\ref{assump:family}, this inverse is well defined on $[0,1]$.
Then, we construct the ironed revenue curve $\overline{REV}_q^{(t)}$ as the upper convex hull of $REV_q^{(t)}$, 
which replaces non-concave regions by linear segments \citep{Myerson1981,ColeRoughgarden2014}.
The resulting curve is concave and preserves the maximum revenue, i.e.,
$\max_{q\in[0,1]} \overline{REV}_q^{(t)}(q) = \max_{q\in[0,1]} REV_q^{(t)}(q) = OPT^{(t)}$.
Thus the optimization problem is globally well posed in quantile space. The set of maximizers forms an interval, which reduces to a single point when the optimum is unique. Each quantile  $q$ in this set corresponds, via $p = F^{(t),-1}(1-q)$, to a revenue-maximizing price, and all such prices yield the same optimal revenue.

\noindent\textbf{Average regret.}
For multiple datasets $D^{(1)},\dots,D^{(T)}$ drawn from the same source $\mathcal{D}$,
the average regret measures the mean revenue shortfall across all sales.
\begin{equation}
\label{eqn:defrt}
    \bar{R}_T = \frac{1}{T}\sum_{t=1}^T r^{(t)}(p^{(t)}).
\end{equation}
Minimizing the average regret requires mechanisms to adapt to valuation uncertainty and refine pricing strategies in sequential settings. While prior work has primarily addressed instantaneous regret for single dataset sales \citep{Segal2003, KleinbergLeighton2003}, this paper addresses the more complex challenge of reducing regret across sequential sales, ensuring long-term revenue optimization.

\section{Mechanism Designs for Data Marketplaces}\label{sec:mapp}

\subsection{MAPP Mechanism}
\label{sec:mappmech}
\noindent
We introduce the MAPP mechanism for each fixed round $t$, which operates in two phases. (i) \emph{Learn via Auction}: The seller conducts a one-time auction with buyers indexed by
$i=1, \dots, n$, referred to as bidders, to gather market information. The auction does not predefine the quantity of datasets to be sold, allowing flexibility in adapting to demand. (ii) \emph{Decide via Posted-Price}: Based on the auction outcomes, the seller sets a uniform price and offers it to all subsequent buyers indexed by $i> n$, referred to as posted-price buyers. We show that this hybrid design ensures fairness by eliminating price discrimination. For simplicity, the round index $t$ is omitted in this subsection. 

\emph{Phase I: Learn via Auction.} 
In this phase, each bidder $i=1,\dots,n$ has a private valuation $v_i$ for the dataset and submits a bid $b_i$, 
yielding bid profile $\bm b = (b_1,\dots,b_n)$.
To ensure bid-independence, each bidder is randomly assigned to a group $\gamma_i\in\{0,1\}$, and 
her price is computed using only bids from the opposite group $\bm b_{[1-\gamma_i]} = \{b_j : \gamma_j = 1-\gamma_i\}$.
We estimate the valuation density with KDE,
\begin{equation}
\label{eqn:defkde}
    \hat f_{\bm b_{[1-\gamma_i]}}(v)
    = \frac{2}{n w} \sum_{b_j\in\bm b_{[1-\gamma_i]}}
        \kappa\!\left(\frac{v-b_j}{w}\right), \qquad v\in[1,H],
\end{equation}
where $\kappa:\mathbb{R}\to\mathbb{R}_+$ is a kernel and $w>0$ a bandwidth.
From $\hat f_{\bm b_{[1-\gamma_i]}}$ we construct the empirical CDF $\hat F_{\bm b_{[1-\gamma_i]}}$, 
the empirical inverse CDF $\hat F_{\bm b_{[1-\gamma_i]}}^{-1}$, 
the empirical revenue curve in quantile space $\widehat{REV}_q$, and 
its ironed version $\overline{\widehat{REV}}_q$  as in Section~\ref{sec:regretmeasure}.

We then select $\hat q_i^* \in \arg\max_{q\in[0,1]} \overline{\widehat{REV}}_q(q)$ and 
define the auction price for bidder $i$ by
\begin{equation}
\label{eqn:defofaucprice}
    p_i := \inf\{p \in [1,H] : 1 - \hat F_{\bm b_{[1-\gamma_i]}}(p) \le \hat q_i^*\}.
\end{equation}
By construction, the auction is bid-independent, 
$h_A(b_i,\bm b_{-i})_i = h_A(b,\bm b_{-i})_i$ for all $b\in\mathbb{R}_+$, 
so each buyer’s price is determined independently of her own bid.
Section~\ref{sec:profincen} shows that MAPP is incentive compatible, so 
bidders optimally bid truthfully and $\bm b$ can be treated as a sample from the valuation distribution.
A bidder purchases the dataset whenever $b_i \ge p_i$.

\emph{Phase II: Decide via Posted-Price.}
After the auction phase, the mechanism transitions to the posted-price phase for buyers who arrive later. The posted price is set as the maximum price determined in the auction phase, ensuring consistency across all buyers:
\begin{equation}
\label{eqn:defofmaxauc}
    p = \max\{p_1, p_2, \dots, p_n\}.
\end{equation}
In this phase, a buyer purchases the dataset if their valuation meets or exceeds the posted price, i.e., $v_i\geq p$ for $i\geq n+1$. The use of a uniform price in Eq.~\eqref{eqn:defofmaxauc} for all remaining buyers ensures fairness in the pricing process.

\subsection{Incentive Guarantees}
\label{sec:profincen}
\noindent
We show that the MAPP mechanism satisfies the incentive measures in Section~\ref{sec:incmeasure}. 
First, it is IR, so buyers do not incur a loss from participating.

\begin{proposition}
\label{prop:mappir}
    The MAPP mechanism satisfies individual rationality.
\end{proposition}
\noindent
The proof of Proposition~\ref{prop:mappir} is provided in Appendix. The key idea is that the posted price in the MAPP mechanism is set as the maximum auction price in Eq.~\eqref{eqn:defofmaxauc}. This design ensures that buyers are always at least as well-off purchasing the dataset during the auction phase as they would be in the posted-price phase. By encouraging participation in the auction phase, buyers have the opportunity to secure the dataset at a potentially lower price, thereby guaranteeing non-negative utility.

Second, MAPP is IC so that buyers bid truthfully to maximize their utility.

\begin{proposition}
\label{prop:mappic}
The MAPP mechanism satisfies incentive compatibility.
\end{proposition}
\noindent
We establish this result by employing a bid-independent auction pricing function $h(\cdot)$ as in Eq.~\eqref{eqn:defofaucprice}, where a buyer’s price is determined based on the bids of others, not their own. This bid-independence prevents buyers from manipulating their prices by adjusting their bids, ensuring that truthful bidding is their utility-maximizing strategy. While bid-independence has been studied in single-phase auctions \citep[e.g.,][]{GoldbergEtAl2006}, the MAPP mechanism extends it to a two-phase structure. By integrating a subsequent posted-price phase informed by the auction outcomes, the mechanism ensures truthful bidding even when buyers have an additional opportunity to purchase, maintaining fairness and efficiency across both phases.

Beyond these two incentive guarantees, MAPP further mitigates price discrimination:
it reduces the first–degree discrimination in \citet{Segal2003} and \citet{KleinbergLeighton2003} 
to third–degree discrimination by restricting prices to at most two levels.
It ensures that identical copies of the same dataset are sold at a single price 
while offering buyers who reveal their valuations in the initial auction a potential compensation. 
This structure fosters predictability and fairness, enhancing buyer trust and encouraging long-term engagement, making MAPP a more stable and sustainable approach for data marketplaces.

\subsection{Online MAPP Mechanism}\label{sec:learning}
\noindent

We now extend the MAPP mechanism in Section~\ref{sec:mappmech} to an online setting 
for selling a sequence of datasets $D^{(1)},\dots,D^{(T)}$ over rounds $t=1,\dots,T$. 
As new datasets are released, the seller iteratively learns optimal pricing strategies by leveraging historical data, 
with each dataset assumed to satisfy Assumption~\ref{assump:family}.
The procedure is given in Algorithm~\ref{algo} and consists of two stages:

(i) \textbf{Exploration stage:}
Apply MAPP with KDE-based value estimates $\hat f_{\mathbf b_{[1-\gamma]}^{(t)}}$ defined in Eq.~\eqref{eqn:defkde} to the datasets $D^{(t)}$ for $t = 1,\dots,\sqrt{T}$. This stage focus on gathering market information by estimating value distributions for these datasets.

(ii) \textbf{Exploitation stage.}
Apply MAPP with RDE-based \citep{QiuEtAl2022} value estimates to datasets $D^{(t)}$, $t=\sqrt{T}+1,\dots,T$, 
using only a small number of bids per round to adjust prices. The RDE is described in detail below.

\SetAlgoNlRelativeSize{-1}
\SetAlCapFnt{\small}
\SetAlCapNameFnt{\small}
\SetAlFnt{\footnotesize}
\begin{algorithm}[htbp]
    \DontPrintSemicolon
    \caption{Online MAPP Mechanism}
    \label{algo}
    \KwIn{Horizon $T$, bids $\bm b^{(t)}\in[1,H]^{n^{(t)}}$ for $t=1,\dots,T$.}
    Set $\tau = \sqrt{T}$.\;
    \For{$t = 1,\dots,T$}{
        Collect bids $\bm b^{(t)}$.\tcp*{Phase I: Learn via auction}
        Draw group assignments $\gamma_i^{(t)}\in\{0,1\}$ for $i=1,\dots,n^{(t)}$.\;
        \For{buyer $i = 1,\dots,n^{(t)}$}{
            \uIf{$t \le \tau$}{
                Estimate $\hat f_{\bm b_{[1-\gamma_i]}^{(t)}}$ via KDE (Eq.~\eqref{eqn:defkde}). \tcp*{Exploration stage} 
            }
            \Else{
                Estimate $\hat f_{\bm b_{[1-\gamma_i]}^{(t)}}$ via RDE (Eq.~\eqref{eqn:defapproxfamily}) 
                with $\hat{\bm\theta}_{\bm b_{[1-\gamma_i]}^{(t)}}$ in Eq.~\eqref{eqn:defthetamle}.\tcp*{Exploitation stage}  
            }
            Construct $\hat F_{\bm b_{[1-\gamma_i]}^{(t)}}$, $\hat F_{\bm b_{[1-\gamma_i]}^{(t)}}^{-1}$,
            $\widehat{REV}_q$, and $\overline{\widehat{REV}}_q$ \linebreak
            as empirical counterparts of the population objects in Section~\ref{sec:regretmeasure}.\;
            Choose $\hat q_i^* \in \arg\max_{q\in[0,1]} \overline{\widehat{REV}}_q(q)$ and 
            set $p_i^{(t)}$ using Eq.~\eqref{eqn:defofaucprice}.\;
            Allocate a copy if $x_i^{(t)} = \mathbf{1}\{b_i^{(t)} \ge p_i^{(t)}\} = 1$.\;
        }
        \For(\tcp*[f]{Phase II: Decide via posted price}){subsequent buyer $i = n^{(t)} + 1, n^{(t)} + 2\dots$}{ 
            Set posted price $p^{(t)}$ according to Eq.~\eqref{eqn:defofmaxauc}.\;
            Allocate a copy if $x_i^{(t)} = \mathbf{1}\{v_i^{(t)} \ge p^{(t)}\} = 1$.\;
        }
    }
\end{algorithm}

In exploitation rounds $t=\tau+1,\dots,T$ with RDE, where $\tau=\sqrt{T}$, we focus on the group that purchases at the lower  auction price in each round, 
because this group determines the Phase II posted price for dataset $D^{(t)}$. 
Let $\tilde{\bm b}^{(t)}$ denote their bids, and $\tilde f^{(t)}$ the corresponding KDE.
We treat $\{\tilde f^{(t)}\}_{t=1}^{\tau}$ as a sample from the shared family and 
apply functional principal component analysis \citep[FPCA,][]{HuynhEtAl2011} 
to their centered log-densities $\{\check f^{(t)}\}_{t=1}^{\tau}$, where
$\check f^{(t)}(v) := \log \tilde f^{(t)}(v) - \frac{1}{H-1}\int_1^H \log \tilde f^{(t)}(s)\,ds$.
This yields an empirical mean function $\hat\mu$ and leading eigenfunctions $\{\hat\varphi_k\}_{k=1}^K$, 
which together define an approximate exponential family 
$\hat{\mathcal{F}} = \{\hat{f}(\cdot|\hat{\bm{\theta}}):\hat{\bm{\theta}}\in\mathbb{R}^K\}$:
\begin{equation}
\label{eqn:defapproxfamily}
    \hat{f}(v|\hat{\bm{\theta}}) = \exp\left(\hat{\mu}(v) + \sum_{k=1}^{K}\hat{\theta}_k\hat{\varphi}_k(v)-\hat{B}_{K}(\hat{\bm{\theta}})\right), \qquad v \in [1,H],
\end{equation}
where $\hat B_K(\theta)$ is the log-normalizing constant. Then the dataset $D^{(t)}$ is priced using this approximating family. 
As in exploration, we draw group assignments $\gamma_i^{(t)}\in\{0,1\}$ for buyers $i=1,\dots,n^{(t)}$.
For bidder $i$, we use bids from the opposite group $\bm b_{[1-\gamma_i]}^{(t)}$ to 
estimate a dataset-specific parameter $\hat{\bm\theta}_{\bm b_{[1-\gamma_i]}^{(t)}}$ by 
maximum likelihood under $\hat f(\cdot\mid\bm\theta)$ in Eq.~\eqref{eqn:defapproxfamily}:
\begin{equation}
\label{eqn:defthetamle}
    \hat{\bm{\theta}}_{\bm{b}_{[1-\gamma_i]}^{(t)}} 
    = \argmax_{\bm\theta:\, \hat{B}_K(\bm{\theta})<\infty} 
      \left\{\sum_{k=1}^K \theta_k \left[\frac{2}{n^{(t)}} 
      \sum_{b_j^{(t)} \in \bm{b}_{[1-\gamma_i]}^{(t)}} 
      \hat{\varphi}_k\bigl(b_j^{(t)}\bigr)\right] - \hat{B}_K(\bm{\theta})\right\}.
\end{equation}
The resulting RDE density $\hat f(\cdot\mid\hat{\bm\theta}_{\bm b_{[1-\gamma_i]}^{(t)}})$
then replaces the KDE in the pricing: auction prices $p_i^{(t)}$ are computed via Eq.~\eqref{eqn:defofaucprice}, 
and the posted price is set as their maximum according to Eq.~\eqref{eqn:defofmaxauc}.
By leveraging historical bid data and the structural patterns estimated during the exploration stage, RDE yields more accurate estimates of the valuation distributions for new datasets.

\section{Regret Analysis}\label{sec:theoretical}
\noindent 
In this section, we provide a regret analysis to evaluate the efficiency of MAPP in optimizing revenue. 
We first study the instantaneous regret when selling a single dataset with MAPP, 
then derive average regret over multiple rounds for the online extension of MAPP.
In both analyses we ignore group splitting and use all $n^{(t)}$ bids to estimate value distribution, 
since splitting merely halves the sample size and does not change the rates of convergence. We introduce some notations.
For sequences $\{a_n\}$ and $\{b_n\}$, we write $a_n = O(b_n)$ if there exist $C > 0$ and $N$ such that
$|a_n| \le C |b_n|$ for all $n > N$, and $a_n \asymp b_n$ if $a_n = O(b_n)$ and $b_n = O(a_n)$. 
For random variables $\{X_n\}$, we write $X_n = O_p(b_n)$ if 
for every $\varepsilon > 0$ there exist $C > 0$ and $N$ with $\Pr(|X_n| > C b_n) < \varepsilon,\,\forall n > N$.
Write $a_n = o(b_n)$ if $a_n/b_n \to 0$.

\subsection{Instantaneous Regret}
\label{sec:instreg}
\noindent
 
For each round $t$, let $q_*^{(t)} \in \arg\max_{q\in[0,1]} \overline{REV}_q^{(t)}(q)$ and 
$p_*^{(t)} := F^{(t),-1}\!\bigl(1 - q_*^{(t)}\bigr)$ 
denote the population optimal sale quantile and its corresponding price, 
so that $OPT^{(t)} = \overline{REV}_q^{(t)}\!\bigl(q_*^{(t)}\bigr)$.
In round $t$ we observe bids $\bm b^{(t)} = \{b_1^{(t)},\dots,b_{n^{(t)}}^{(t)}\}$ and
form an estimator $\hat f^{(t)}$ of the value density, which 
induces an empirical CDF $\hat F^{(t)}$, its generalized inverse $\hat F^{(t),-1}$, and 
an empirical ironed revenue curve $\overline{\widehat{REV}}_q^{(t)}$ as 
in Section~\ref{sec:regretmeasure}. 
We then set $\hat q_*^{(t)} \in \arg\max_{q\in[0,1]} \overline{\widehat{REV}}_q^{(t)}(q)$ and 
$\hat p_*^{(t)} := \hat F^{(t),-1}\!\bigl(1 - \hat q_*^{(t)}\bigr)$,
which serve as the empirical optimal sale quantile and price in the instantaneous-regret analysis.

\begin{proposition}[Instantaneous regret bound]
\label{prop:instreg}
   Under Assumption~\ref{assump:family} and the interior condition $p_*^{(t)} \in (1,H)$ for all $t$, 
let $f^{(t)}$ denote the true valuation density and $\hat f^{(t)}$ its estimator.
Then the instantaneous regret in Eq.~\eqref{eqn:definstreg}  satisfies $r^{(t)}(\hat p_*^{(t)}) = O\big(\|\hat f^{(t)} - f^{(t)}\|_\infty^2\big)$.
\end{proposition}

\noindent
This proposition shows that instantaneous regret is controlled by the squared uniform density estimation error. 
Building on this, we next analyze the instantaneous regret of online MAPP in Algorithm~\ref{algo}, 
separately for the exploration and exploitation stages.

\subsubsection{Exploration Stage}
\label{sec:explore}
\noindent
In this stage, the value density is estimated via the KDE in Eq.~\eqref{eqn:defkde}, 
where the kernel $\kappa$ satisfies the usual regularity conditions,
for constants $\rho, C_\rho, s_0>0$,
\begin{equation}
\label{eqn:condkde}
\begin{aligned}
    & \int_{-1}^1 \kappa(u)du = 1 \text { (validity);} 
    && \kappa(u) = \kappa(-u)\ \forall u\in[-1, 1] \text { (symmetry);}\\
    & \int_{-1}^1 \kappa(u)^2du<\infty \text{ (finite $L_2$);}
    && \int_{-1}^1 u^2\kappa(u)du<\infty \text { (finite variance);} \\
    & \int_{-1}^1 u\kappa(u)du = 0 \text{ (zero mean);}
    && \kappa(|s|)\leq C_\rho\cdot\exp(-s^\rho) \ \forall s>s_0 \text{ (exponential decay).}
\end{aligned}
\end{equation}

These regularity conditions are satisfied by common kernel choices used in practice \citep[e.g.,][]{Parzen1962}. 
Any boundary bias near $1$ or $H$ can be handled by standard boundary-corrected KDE on $[1,H]$.

\begin{theorem}[Exploration-stage KDE regret bound]
\label{thm:kde}
Under Assumption~\ref{assump:family} and the kernel conditions in Eq.~\eqref{eqn:condkde}, 
suppose the exploration-stage KDE in Eq.~\eqref{eqn:defkde} uses the bandwidth $w = O\!\left(n^{-1/3}\right)$, 
where $n$ is the number of bids received in the current round.
Then, for all $n>1$, with probability at least $1-1/n$, the instantaneous regret in the current round satisfies
$r \;\le\; C_{\mathrm{KDE}}\, n^{-2/3}\log n$,
for some constant $C_{\mathrm{KDE}} > 0$ independent of $n$.
\end{theorem}

\noindent 
Compared to the KDE rate in \citet{Segal2003}, our approach relaxes its underlying assumptions. 
Specifically, it does not require 
the true density $f$ to be continuously differentiable with a Hölder continuous derivative, 
nor does it impose Hölder continuity on the kernel function. 
Additionally, our result provides a finite-sample guarantee, 
making it practical for data marketplaces where $n$ may be limited.


\subsubsection{Exploitation Stage}
\label{sec:exploit}
\noindent
In this stage, KDE is replaced with RDE, using exploration-stage bids to refine the density estimate. 
For identifiability and stability of RDE, we impose the following additional regularity on $\mathcal F$, 
in addition to Assumption~\ref{assump:family} and the kernel conditions in Eq.~\eqref{eqn:condkde}.

\begin{assumption}[Functional structure of the shared family]
\label{assump:regularfamily}
(i) \textbf{Shape-regular densities.}
All $f^{(t)}$ are continuously differentiable on $[1,H]$, with 
$|(f^{(t)})'|$ bounded above and away from zero on $[1,H]$, uniformly in $t$.
(ii) \textbf{FPCA structure.}
The centered log-densities of $f^{(t)}$ admit a well-defined FPCA eigenstructure:
for some fixed $K$, the first $K$ eigenvalues $\{\lambda_k\}_{k=1}^K$ are strictly positive and well separated, 
the corresponding eigenfunctions $\{\varphi_k\}_{k=1}^K$ are continuously differentiable with 
$|\varphi_k'|$ bounded above and away from zero on $[1,H]$, and the remaining spectral tail is negligible. 
The FPCA computed on the estimated log-densities $\log \hat f^{(t)}$ yields 
leading eigenfunctions sign-aligned with their population counterparts.
\end{assumption}

\noindent
The shape-regular condition imposes mild smoothness on the densities that rules out flat regions or sharp turning points.
The FPCA condition is a standard assumption 
ensuring a stable, identifiable low-dimensional structure of the shared valuation density family.
Under the framework of \citet{QiuEtAl2022}, RDE provides $L_1$ bounds on the density estimation error. 
Proposition~\ref{prop:instreg} converts these statistical guarantees into regret guarantees, 
bur requires a uniform ($L_\infty$) bound. Bridging this gap is a key technical contribution of this paper.
The next proposition establishes this bridge, 
upgrading $L_1$ control to $L_\infty$ under the RDE regularity assumptions and 
the design conditions specified below.

Recall that $\tau$ denotes the number of exploration rounds.
Let $n$ be the number of bids observed in the current exploitation round, 
$m := \min_{1 \le t \le \tau} n^{(t)}$ the minimal number of bids observed in any exploration round, 
$w$ the KDE bandwidth, 
$K(\tau)$ the FPCA truncation level, and 
$\{\lambda_k\}_{k\ge1}$ the eigenvalues in Assumption~\ref{assump:regularfamily}.  
The bandwidth $w \to 0$ satisfies 
\begin{equation}
\label{eqn:rde-growth}
    w \asymp m^{-1/3}, \quad
    |\log w| = O(\log \tau), \quad
    \frac{\log \tau}{m} = o(w), \quad
    \frac{w\,(w + 1/m)\,\log \tau}{\tau} \to 0.
\end{equation}
We regard $K(\tau)$ as a tuning parameter that can be selected to control the truncation bias.
In particular, we assume it is chosen so that the eigenvalue tail is negligible at the target rate.
For the RDE asymptotic regime, corresponding to many exploration auctions and a growing exploitation market,
we consider the design
\begin{equation}
\label{eqn:rde-design}
  \tau \asymp n, \quad
  m \asymp n^{3/2}, \quad
  K(\tau) = O(\log \tau), \quad
  \sum_{k > K(\tau)} \lambda_k = O(\tau^{-1}).
\end{equation}

\begin{proposition}[From $L_1$ to $L_\infty$ error for RDE]
\label{prop:L1-Linfty}
Let $f^{(t)}$ and $\hat f^{(t)}$ denote the true valuation density and its RDE estimator in Eqs.~\eqref{eqn:defapproxfamily}-\eqref{eqn:defthetamle}. 
Under Assumptions~\ref{assump:family}-\ref{assump:regularfamily}, 
suppose the RDE employs the KDE estimator Eq.~\eqref{eqn:defkde} with 
kernel satisfying Eq.~\eqref{eqn:condkde} and 
bandwidth satisfying Eq.~\eqref{eqn:rde-growth}, 
under the RDE design specified in Eq.~\eqref{eqn:rde-design}.  
Then there exists a constant $C>0$, independent of $t$, such that, with high probability,
$\bigl\|\hat f^{(t)} - f^{(t)}\bigr\|_\infty 
    \;\le\; C\,\bigl\|\hat f^{(t)} - f^{(t)}\bigr\|_1$.
\end{proposition}

\noindent
Based on this result  of RDE and the regret bound in Proposition~\ref{prop:instreg}, we establish instantaneous regret bound for RDE. 

\begin{theorem}[Exploitation–stage RDE regret bound]
\label{thm:rde}
Under the conditions of Proposition~\ref{prop:L1-Linfty}, for all $n > 1$,
there exists a constant $C_{\mathrm{RDE}} > 0$, independent of $n$ and $t$, such that 
with probability at least $1 - 1/n$, the instantaneous regret in round $t$ in Eq.~\eqref{eqn:definstreg} satisfies
$r^{(t)} \;\le\; C_{\mathrm{RDE}}\,n^{-1}(\log n)^2$.
\end{theorem}

\noindent
MAPP with RDE-based estimates achieves an instantaneous regret rate of $O_p\!\big(n^{-1}(\log n)^2\big)$, 
improving on the classical $O_p\!\big(n^{-2/3}(\log n)^{2/3}\big)$ bounds for dynamic posted pricing
\citep{Segal2003, KleinbergLeighton2003}.
These dynamic posted-price mechanisms are the only existing framework that, like MAPP,
accommodates an unbounded stream of arriving buyers and is therefore directly comparable in terms of realized revenue.
The statistical engine underlying our guarantees is the RDE
of \citet{QiuEtAl2022}, which we use as an off-the-shelf density estimator.
Our main technical contribution is to relate the convergence properties of the density estimator to revenue performance in Propositions~\ref{prop:instreg} and \ref{prop:L1-Linfty}. 

Moreover, under the Lipschitz condition in Assumption~\ref{assump:family}, the minimax lower bound for estimating $|\hat f^{(t)} - f^{(t)}|_\infty^2$ is of order $O_p(n^{-2/3})$, and under stronger parametric assumptions it improves to $O_p(n^{-1})$ \citep{Tsybakov2008Introduction}. Therefore, Propositions~\ref{prop:instreg} and \ref{prop:L1-Linfty} imply that the regret bounds in Theorems~\ref{thm:kde} and \ref{thm:rde} are rate-optimal up to logarithmic factors.


\subsection{Average Regret}
\label{sec:cumreg}
\noindent
We now aggregate the instantaneous regret bounds to obtain the average regret of online MAPP across all $T$ rounds, 
as defined in Eq.~\eqref{eqn:defrt}. 
In our marketplace, each round $t$ corresponds to selling a new dataset: 
the seller first runs an auction to elicit bids for that dataset and then 
sets a posted price for the same round.
We split the horizon at an exploration length $\tau$: 
rounds $1,\dots,\tau$ are used for exploration, and rounds $\tau+1,\dots,T$ for exploitation.
Let $m_{\mathrm{explore}} := \min_{1 \le t \le \tau} n^{(t)}$ and 
$m_{\mathrm{exploit}} := \min_{\tau < t \le T} n^{(t)}$ denote 
the minimum per–round numbers of bids in the exploration and exploitation stages, respectively.
The next theorem specifies an online design for $(\tau, m_{\mathrm{explore}}, m_{\mathrm{exploit}})$ and 
the resulting average regret rate.

\begin{theorem}[Average regret bound for online MAPP]
\label{thm:regret}
Under the same regularity, kernel, and RDE-design conditions as Proposition~\ref{prop:L1-Linfty}, 
assume that, as $T \to \infty$,
$\tau \asymp T^{1/2}, m_{\mathrm{explore}} \asymp T^{3/4}(\log T)^{-2},$ and $m_{\mathrm{exploit}} \asymp T^{1/2}(\log T)^{-2}$.
Then the average regret in Eq.~\eqref{eqn:defrt} satisfies
$\bar{R}_T = O_p\bigl(T^{-1/2}(\log T)^2\bigr)$.
\end{theorem}

\noindent
This theorem  shows that the average regret of MAPP converges to zero, so the long-run revenue of the seller
converges to that of an oracle with full knowledge of each dataset’s valuation distribution. In particular, 
MAPP is Hannan consistent \citep{Hannan1957}, satisfying $\limsup_{T\to\infty}\bar R_T = 0$. 

This result provides, to our knowledge, the first regret guarantee for a setting in which heterogeneous goods, each available in unlimited copies, are sold to buyers who arrive and depart dynamically. This represents a more challenging learning problem than the traditional repeated-sales model, where a single product with a fixed valuation distribution is sold repeatedly and prices can be refined through accumulating observations
\citep[see,][]{BergemannSaid2011, CesaBianchiEtAl2015}.
In our setting, each dataset arrives with its own valuation distribution. 
Despite this additional complexity, the average regret rate we obtain matches the best-known results for the simpler single-product problem \citep{GuoEtAl2022, ChenEtAl2023}, and aligns with the minimax lower bound for that setting up to logarithmic factors \citep{beygelzimer2011contextual}.

\section{Numerical Experiments}\label{sec:empirical}

\noindent 
In this section, we present numerical experiments to evaluate the effectiveness of the MAPP mechanism.
All MAPP variants share the same pricing rule in Eq.~\eqref{eqn:defofaucprice}, 
differing only in how the CDF term $\hat F_{\bm b_{[1-\gamma_i]}}(p)$ is estimated 
from the bids $\bm b_{[1-\gamma_i]}$, $\gamma_i\in\{0, 1\}$, as described below:  
(1) \textbf{eCDF–MAPP}, which uses the empirical distribution function 
\citep[e.g.,][]{Segal2003, GoldbergEtAl2006, ColeRoughgarden2014},
\begin{equation}
\label{eqn:ecdf}
  \hat F_{\bm b_{[1-\gamma_i]}}(p)
  := \frac{1}{|\bm b_{[1-\gamma_i]}|}
     \sum_{b_j \in \bm b_{[1-\gamma_i]}} \mathbf{1}\{b_j \le p\};
\end{equation}
(2) \textbf{KDE–MAPP}, which estimates the density via KDE (Eq.~\eqref{eqn:defkde}) and 
obtains the CDF by integration; and 
(3) \textbf{RDE–MAPP}, which estimates the density via RDE 
(Eq.~\eqref{eqn:defapproxfamily}–Eq.~\eqref{eqn:defthetamle}) 
and likewise computes the CDF by integration. 
To evaluate how the amount of historical data influences performance, 
we consider three training configurations: 
20 rounds × 200 bids, 200 rounds × 20 bids, and 200 rounds × 200 bids.

To compare MAPP against established auction-based designs, 
we also implement the \textbf{empirical Myerson} auction \citep{ColeRoughgarden2014} and 
the deep-learning-based \textbf{MyersonNet} \citep{DuettingEtAl2021}, trained on 200 rounds × 200 bids.
Both implement the Myerson optimal auction for a single indivisible item, 
driven by the virtual valuation, 
  $\nu(v) := v - \frac{1 - F(v)}{f(v)}$,
for a valuation distribution with CDF $F$ and density $f$, and allocate the item to the bidder
with the highest nonnegative virtual value at the minimum bid needed to retain that position.
The \textbf{empirical Myerson} auction, like eCDF–MAPP, uses the empirical CDF in Eq.~\eqref{eqn:ecdf}
to build and iron an empirical revenue curve, but then runs the induced Myerson auction and
restricts candidate prices to the observed bids.
\textbf{MyersonNet} instead parameterizes the allocation rule $g_\eta(b)$ and payment rule
$t_\eta(b)$ with neural networks, and learns $\eta$ by maximizing expected revenue,
$\mathrm{rev}(g_\eta, t_\eta)
  = \mathbb{E}_{v \sim F}\Bigl[\sum_{i=1}^n g_{\eta,i}(v)\, t_\eta(v)\Bigr]$,
approximated by empirical averages over sampled bid profiles.  
In effect, MyersonNet learns virtual valuation functions that maximize empirical auction revenue.

Both the empirical Myerson auction and MyersonNet are designed for a single-item setting, 
where at most one bidder receives the good, whereas in our data marketplace the seller can supply 
the same dataset to essentially unlimited buyers in a round.  
To place these Myerson-based mechanisms on equal footing for comparison, 
we instead take the reserve price that makes the virtual valuation zero \citep{Myerson1981} and 
treat this as the posted price offered to all buyers.  
This adaptation does not make them tailored to our setting, but 
it brings their revenues to the same scale as MAPP, allowing a consistent regret comparison. 

We evaluate mechanism performance using the average regret in Eq.~\eqref{eqn:defrt}, 
based on 1,000 independent experiments of 1,000 auction rounds, 
which yields one average regret value per mechanism in each experiment. 
In the simulated environments, 
valuations are drawn from distributions supported on $[1,10]$ as in Assumption~\ref{assump:family}. 
For the real bidding data, we linearly rescale observed bids to the same support $[1,10]$ before resampling. 
To study market thickness, 
we repeat the experiments under four fixed numbers of bidders per round (10, 50, 100, and 200). 
For each valuation specification, all mechanisms are evaluated on the same valuation samples within an experiment, 
so differences in outcomes reflect the mechanisms rather than sampling variability. 
Figures~\ref{fig:regret-sim} and~\ref{fig:regret-real} show the resulting histograms of average regrets 
for the simulated and real bidding data. Across all configurations, 
RDE–MAPP achieves the lowest and least variable regret among the mechanisms considered.


\subsection{Simulated Data}
\label{sec:sim}

\noindent
In the simulated data experiments, we instantiate the shared valuation family in Assumption~\ref{assump:family} 
using four standard parametric specifications from mechanism design and auction theory:
(1) truncated normal, (2) beta, (3) truncated exponential, and (4) truncated Pareto. 
Across rounds, the parameters of the valuation density may change, 
but the density always remains within the same chosen family.

\begin{figure}[htb]
    \centering
    \includegraphics[width=0.85\textwidth]{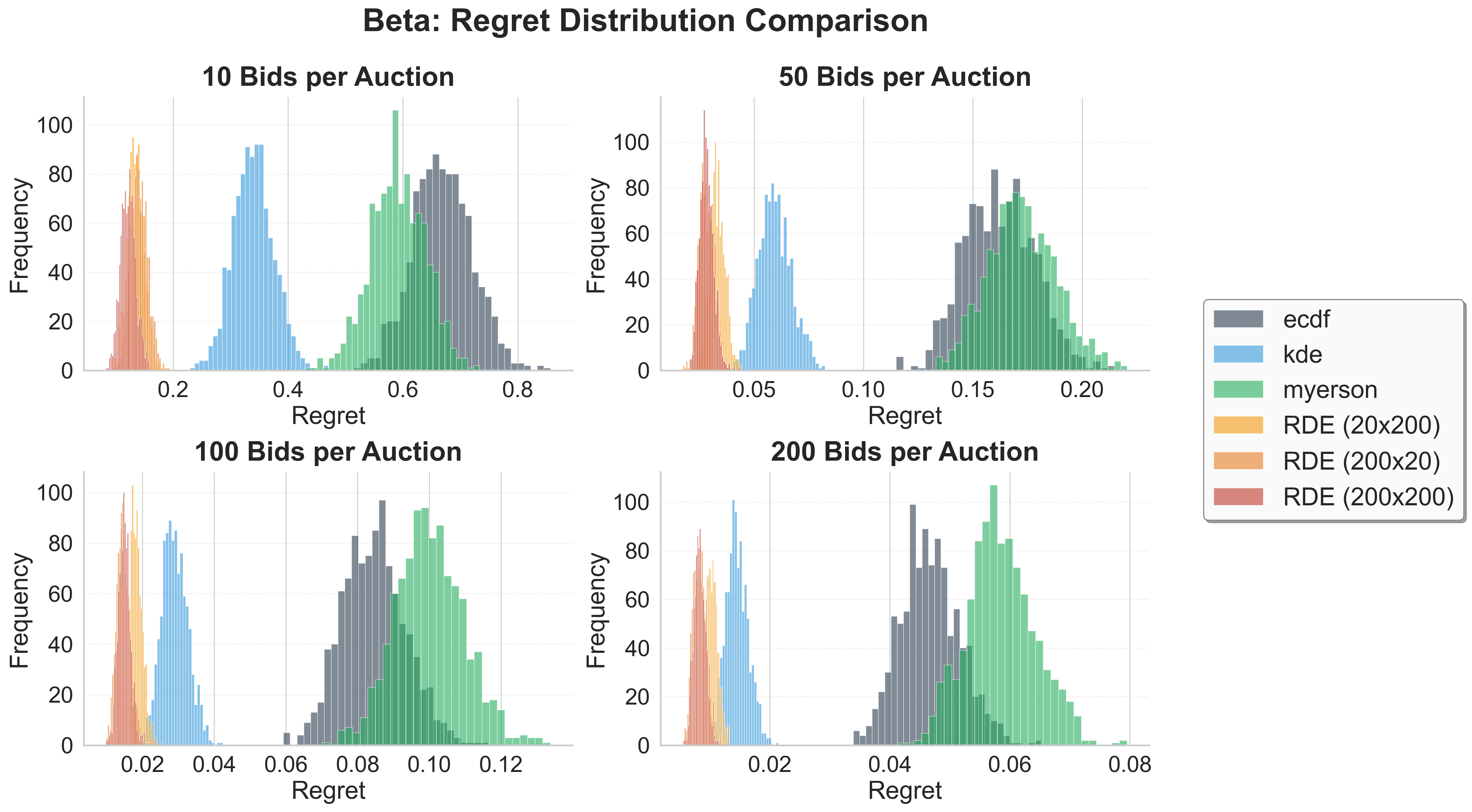}
    \caption{Histogram of 1,000 average regret values for each mechanism 
    under the simulated beta valuation specification. 
    Each value is computed from one independent experiment of 1,000 auction rounds. 
    Lower values indicate better performance.}
    \label{fig:regret-sim}
\end{figure}

For each family, we restrict parameters to exclude extremely flat or extremely heavy-tailed cases 
that would require substantially more training rounds and much larger sample sizes per round to estimate reliably
(see Appendix  for specific ranges). 
These restrictions are imposed to keep computation time at a reasonable level on standard desktop hardware, 
rather than because of any inherent limitation of the mechanisms. 
Under these choices, the truncated normal specification represents balanced markets with mostly medium-valuation buyers; 
the beta family represents skewed markets with demand tilted toward lower or higher valuations; 
the truncated exponential family models markets with many low-valuation buyers 
and rapidly thinning high-valuation demand; 
and the truncated Pareto family corresponds to heavy-tailed markets 
where many buyers have moderate valuations and a few exhibit occasionally very high valuations. 
Figure~\ref{fig:regret-sim} displays a representative histogram of average regret under the beta specification. 
Histograms for the truncated normal, truncated exponential, and truncated Pareto specifications are deferred to Appendix and show similar patterns.

In the simulated environments, RDE–MAPP yields 
the leftmost and most concentrated distribution of average regret across all configurations.
Among the three RDE training configurations, 
the setting with 200 training rounds and 200 bids per round performs best overall, 
which is consistent with the expectation that larger training samples lead to more accurate density estimates. 
KDE-based pricing is competitive only in the thickest markets we consider, with 200 bidders per auction. 
Mechanisms based on the empirical CDF exhibit substantially higher regret across all bidder counts, 
with the empirical Myerson auction performing worse than the corresponding empirical CDF. 
MyersonNet performs even worse, with its regret distribution lying far to the right of the other methods; its detailed results are reported in the Appendix.

\subsection{Real Data}
\label{sec:real}

\noindent
To evaluate the practical applicability of our mechanism, 
we use bid records from the FCC AWS-3 spectrum auction \citep{Milgrom2004}
\footnote{Data and details: https://www.fcc.gov/auction/97.}. 
Although spectrum licenses are finite-supply goods, this auction provides 
one of the few publicly available datasets with large-scale bidding on many items of similar type,
yielding empirical bid distributions across license categories that are broadly similar and 
thus consistent with Assumption~\ref{assump:family}.

The AWS-3 auction used a simultaneous multiple-round ascending format, 
in which bidders may repeatedly bid on a license until exiting, and 
final bids are widely used as proxies for private valuations \citep{Cramton1997,BajariFox2005}.
It concluded after 341 rounds and involved 70 bidders competing for 1,614 licenses of six types (A1, B1, G, H, I, J). 
Within each type, licenses differ only by geographic region, 
so a license type can be viewed as an analogue of a dataset sold repeatedly in our setting. 
For each type, we use the empirical distribution of exit bids to approximate its underlying valuation distribution.

\begin{table}[htb]
    \centering
    \footnotesize
    \begin{tabular}{lccc}
        \hline
        \textbf{Type} & \textbf{Number of Bids} & \textbf{Min of $\log$(Bid)} & \textbf{Max of $\log$(Bid)} \\
        \hline
        A1 & 679 & 7.24 & 16.54 \\
        B1 & 846 & 7.60 & 19.74\\
        G & 4286 & 6.91 & 20.54 \\
        H & 1126 & 9.62 & 20.97 \\
        I & 1037 & 9.95 & 21.00 \\
        J & 923 & 10.76 & 21.74 \\
        \hline
    \end{tabular}
    \caption{Summary statistics for FCC AWS-3 auction data.}
    \label{tab:summary}
\end{table}

\begin{figure}[htb]
    \centering
    \begin{subfigure}[t]{0.43\textwidth}
        \centering
        \includegraphics[width=\textwidth, height=0.7\textwidth]{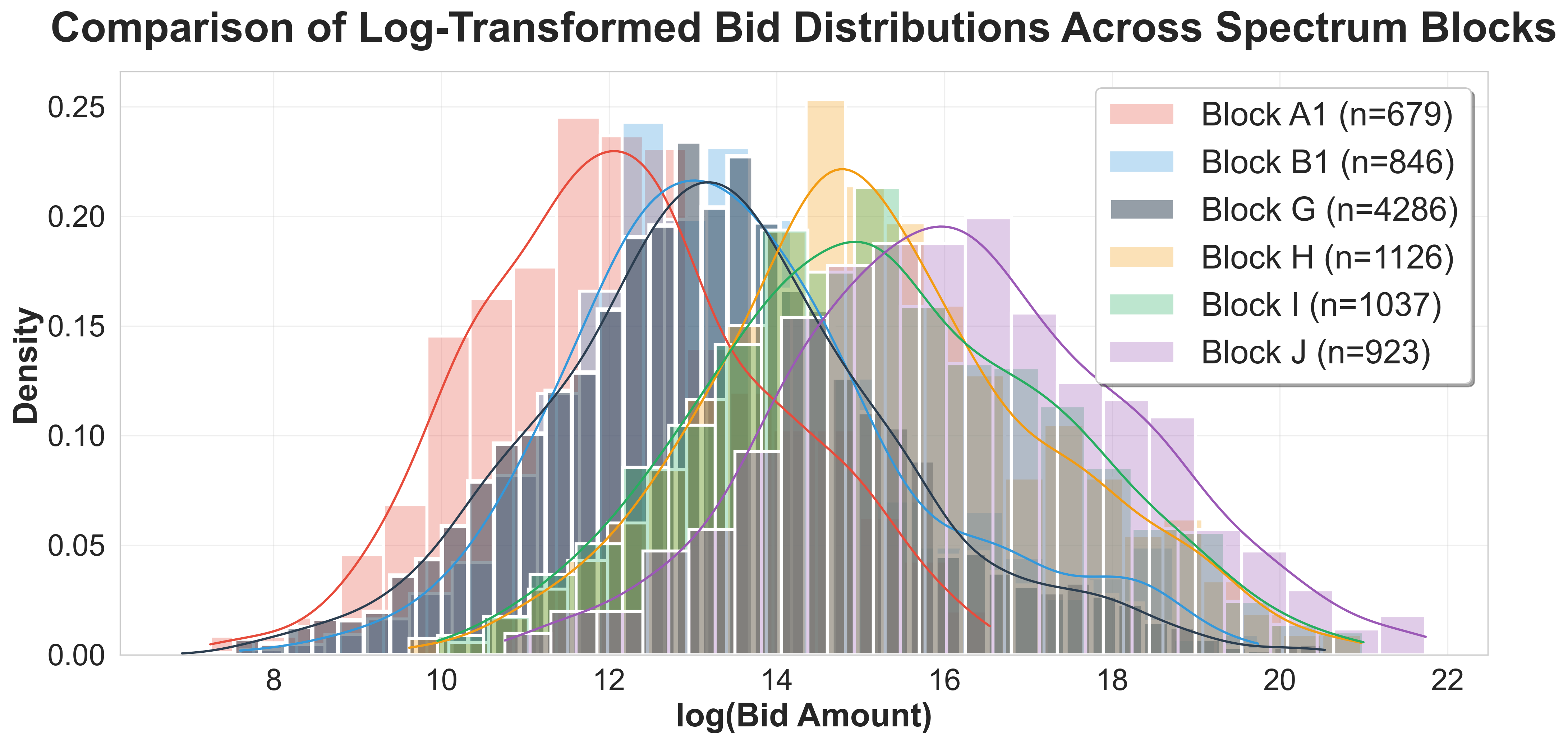}
    \end{subfigure}
    \quad
    \begin{subfigure}[t]{0.43\textwidth}
        \centering
        \includegraphics[width=\textwidth, height=0.7\textwidth]{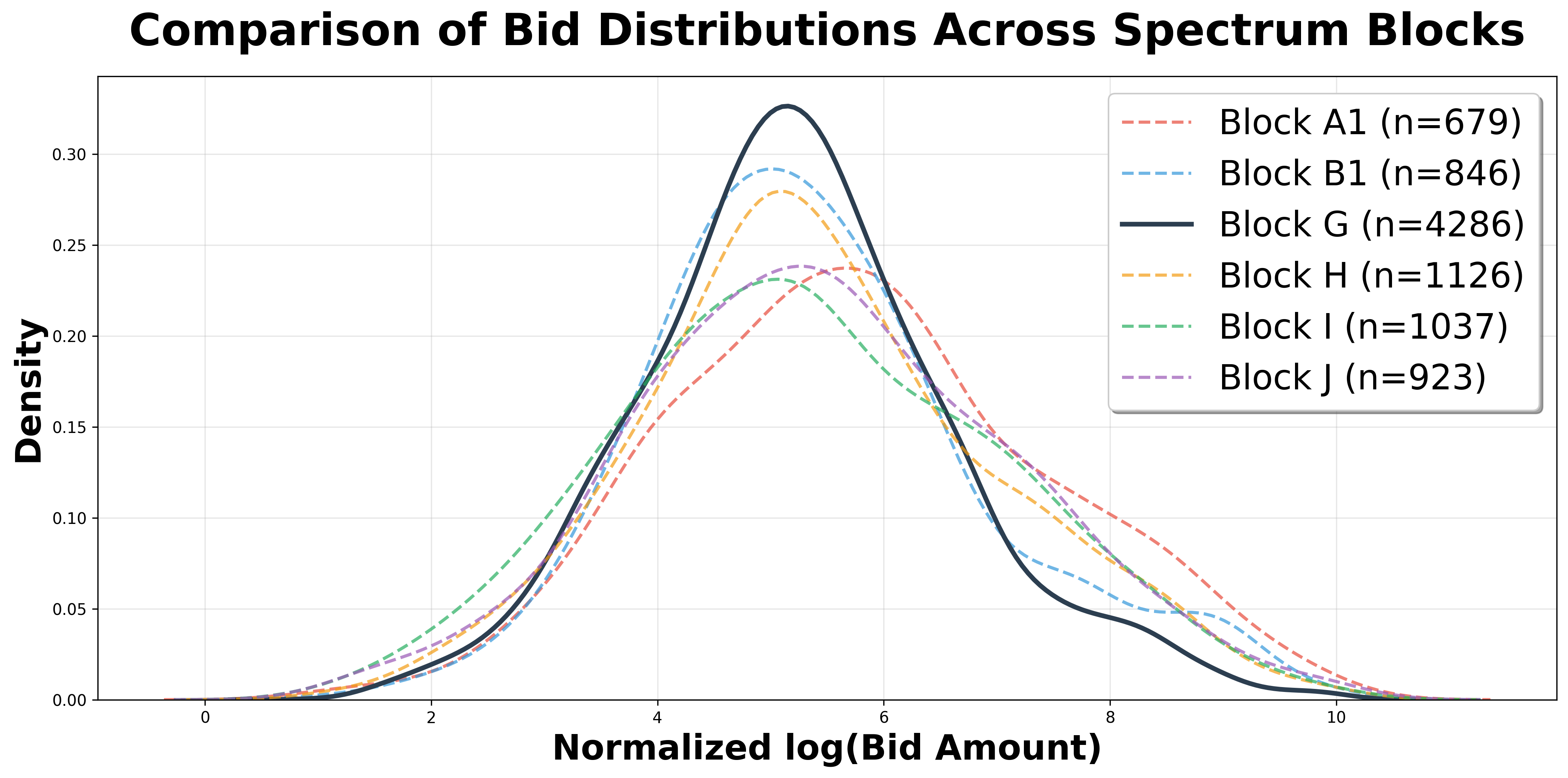}
    \end{subfigure}
    \caption{Empirical valuation distributions from AWS-3 auction.
    The left panel shows histograms with kernel density estimates for the log-transformed bids across six license types. 
    The right panel shows kernel density estimates for the corresponding rescaled log-transformed bids.}
    \label{fig:real}
\end{figure}

Raw bids are strongly right-skewed, so we first apply a logarithmic transformation and then 
linearly rescale the transformed bids to the common support $[1,10]$. 
Table~\ref{tab:summary} summarizes the log-transformed bids before rescaling by license type, 
reporting the number of bids and the minimum and maximum values in each type. 
There are ample bids in every type to support resampling in the experiments, and 
the log-transformed bids lie in a moderate range (approximately 6 to 22), 
so the subsequent linear rescaling to $[1,10]$ is a mild normalization rather than a substantial distortion. 
We use license type G as the test distribution and the remaining types as training distributions for RDE. 
Figure~\ref{fig:real} displays the corresponding valuation distributions after log transformation and after rescaling. 
The rescaled distributions have broadly similar shapes, 
providing empirical support for modeling them as draws from a shared density family as in Assumption~\ref{assump:family}.

\begin{figure}[htb]
    \centering
    \includegraphics[width=0.85\textwidth]{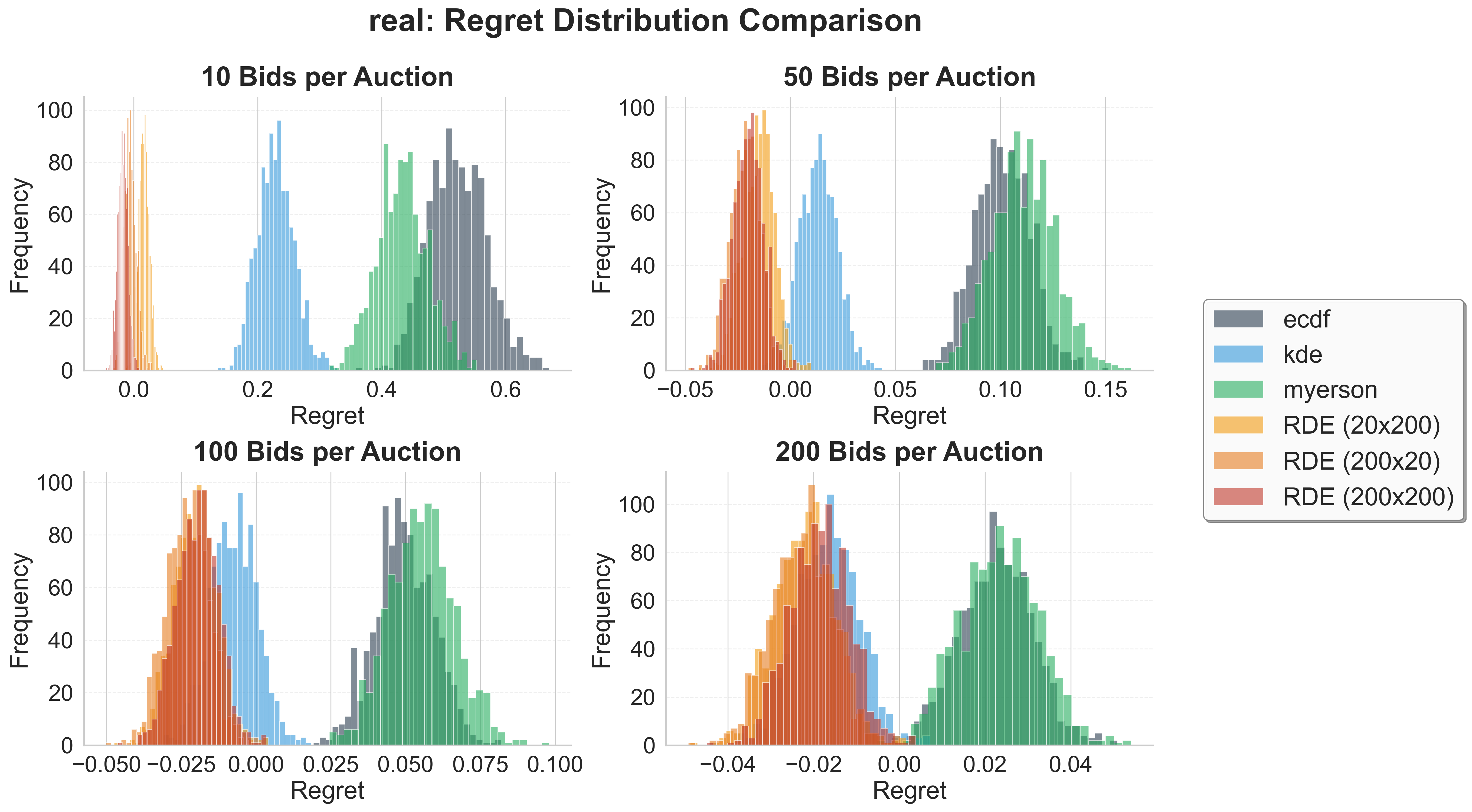}
    \caption{Histogram of 1,000 average regret values for each mechanism 
    under the AWS\textendash 3 valuation specification. 
    Each value is computed from one independent experiment of 1,000 auction rounds, 
    and lower values indicate better performance. }
    \label{fig:regret-real}
\end{figure}

Figure~\ref{fig:regret-real} reports the histograms of average regret for the real bidding data. 
The results closely match the simulated study: 
RDE–MAPP has the lowest mean regret and the tightest dispersion, 
and therefore outperforms the other mechanisms in both level and variability. 
As the number of bidders per round increases, 
the KDE-based pricing rule becomes more competitive and the advantage of RDE narrows, 
which is natural because RDE can use training information only from the five auxiliary license types, 
each with a moderate number of bids. 
For our data marketplace setting, performance in thinner markets with fewer bids is particularly important, 
and in this regime the gains from RDE are most pronounced. 
The remaining mechanisms perform noticeably worse: KDE still lags behind RDE when bids are limited (10 or 50), 
mechanisms based on the empirical CDF and the empirical Myerson auction exhibit substantially higher regret, and MyersonNet is worse still, with its regret distribution far to the right of the others. 
Including MyersonNet in the main histogram would compress the remaining distributions and hinder comparison, 
so versions that include MyersonNet are reported only in Appendix.

Hence, both the simulation and real data results show that RDE–MAPP consistently attains the lowest regret, with the smallest uncertainty, across a broad range of valuation environments and market sizes.
KDE-based pricing approaches this performance only when many bids are available, 
whereas RDE–MAPP maintains a clear advantage when bids per round are limited, 
which is the setting most relevant for data marketplaces. 
Moreover, the sensitivity analysis in Appendix  shows that 
RDE–MAPP is robust to the choice of bidder group splitting, 
so a simple two-group design used in Section~\ref{sec:mappmech} suffices. 
This reduces price discrimination across groups 
while maintaining the revenue performance that makes RDE–MAPP competitive with, and often superior to, KDE.

\section{Related Work}\label{sec:related}
\noindent We briefly review some of the related work from multiple literatures, including mechanism design, data marketplace, and online learning.

\medskip\noindent\textbf{Mechanism Design.}
The sale of digital goods like music, movies, and datasets presents unique mechanism design challenges, as they are non-rivalrous and can be resold indefinitely without depleting inventory. Traditional mechanisms, such as sequential posted pricing \citep{BarYossefEtAl2002} and digital goods auctions \citep{GoldbergEtAl2006}, aimed to fully exploit individual utilities, but this goal proved overly ambitious. Sequential posted pricing can incur arbitrarily bad revenue early on unless bids are concentrated around the optimal price, while digital goods auctions lead to significant revenue losses in worst-case scenarios due to the inability to directly set prices based on buyers’ true valuations. \citet{GoldbergEtAl2006} also showed that individual prices offer no advantage over randomized group pricing, the latter of which is thus incorporated into our MAPP mechanism. 
Later works, such as \citet{Segal2003} and \citet{KleinbergLeighton2003} reintroduced distributional assumptions from \citet{Myerson1981}, while  \citet{BroderRusmevichientong2012} imposed parametric assumptions, all in an effort to address these limitations. These works focus on a single dataset sale and heavily rely on abundant value observations ($n$), with \citet{HuangEtAl2015} discussing the required size of $n$ for near-optimal revenue. In contrast, our MAPP mechanism remains nonparametric and accelerates valuation learning by leveraging historical value observations from similar datasets. It achieves an $O_p(n^{-1/3})$ improvement in regret convergence and overcomes the auction's limitation of waiting indefinitely for all buyers who dynamically enter and exit the market, thereby boosting long-term revenue.

\medskip\noindent\textbf{Data Marketplace.}
The rise of data as a key economic asset has spurred extensive research on designing efficient data marketplaces \citep{BergemannBonatti2019, Pei2022}.
Many works leverage data divisibility to enable customization, often leading to price discrimination. For consumer data, customization helps buyers target specific segments, aligning data with their needs and increasing perceived value \citep{MehtaEtAl2019, Yang2022}. However, for general datasets, where buyers cannot assess value ex-ante, such approaches are less effective \citep{ChenEtAl2022}. To bypass valuation uncertainty, sellers often offer data derivatives like pre-trained models, easing the buyers’ evaluation but obscuring price discrimination \citep{AgarwalEtAl2019}. In contrast, our mechanism reduces price discrimination transparently while maintaining revenue efficiency, making it ideal for scalable and equitable data marketplace pricing.

\medskip\noindent\textbf{Online Learning.}
The integration of online learning with markets has gained traction in recent years \citep{DaiJordan2021NeurIPS, DaiJordan2021JMLR}. When applied to pricing goods with an unlimited supply, it typically focuses on selling identical copies through sequential posted prices \citep{BarYossefEtAl2002, KleinbergLeighton2003} or repeated second-price auctions, where algorithms are often designed to learn the optimal reserve price \citep{NingEtAl2021, KanoriaNazerzadeh2021, CesaBianchiEtAl2015}. However, learning the reserve price from bids is challenging since those below this threshold remain unobserved, restricting the auctioneer’s ability to refine pricing strategies \citep{ChawlaEtAl2014}. Another application of online learning is from the buyer’s perspective, where algorithms are designed to optimize bidding or purchasing strategies \citep{GuoEtAl2022}.
Our work is among the first to apply online learning to the sequential sale of multiple datasets, each with unlimited copies, extending learning both within and across datasets with varying value distributions. This advances online learning in pricing beyond existing frameworks that focus on a single evolving product, demonstrating its applicability to dynamic data marketplaces.

\section{Conclusion}\label{sec:conclusion}

\noindent
This paper proposes the MAPP mechanism, 
which connects auction-based demand learning with posted-price selling in emerging data marketplaces. 
MAPP first runs a bid-independent auction to learn the valuation distribution and then 
sets a posted price for later buyers equal to the maximum auction price. 
Because the auction is bid-independent, truthful bidding is a best response, and 
setting the posted price at least as high as the auction prices ensures individual rationality, 
since buyers do not gain by waiting and only accept the posted price when their utility is nonnegative. 
By using the minimum two bidder groups required for bid independence, 
MAPP keeps price discrimination as low as possible.
We show that revenue optimization in repeated dataset sales can be reframed as a valuation density estimation problem. 
We prove a general regret-density connection under our setup, establishing that  controlling revenue regret reduces to accurately estimating the valuation density. 
MAPP achieves a regret $O_p(n^{-1}(\log n)^2)$ 
for a single dataset with $n$ bids, while 
an online version that sells $T$ datasets in sequence attains regret of order $O_p(T^{-1/2} (\log T)^2)$.
Our simulation studies and a real-data study with the AWS-3 spectrum license market
confirm that MAPP attains the lowest and stable regret compared to alternatives.

There are several promising directions for future work. 
First, one can further exploit the initial learning auction by modeling its transaction and operational costs and adjusting its design or frequency.
Second, it would be useful to study how different signaling policies influence buyers’ valuations, 
comparing public signals with private or personalized ones while maintaining fairness and transparency.
Third, MAPP could be extended to settings with nontrivial costs of maintaining a data delivering infrastructure, 
where serving a new buyer is substantially more expensive than serving an existing one, 
so auctions can be used to offer discounts to frequent buyers while posted prices serve occasional buyers.


\section*{Supplementary Materials}
\noindent
The online supplementary materials include the code for reproducing all results, the technical proofs, and additional numerical studies.

\section*{Acknowledgments}
\noindent
We thank the editor, the associate editor, and the four anonymous reviewers for their valuable comments that helped improve this paper.

\section*{Funding}
\noindent
This research was partially funded by grants from the NIH (R01DK142026, HZ, JZ, and XD; R35GM141798, HZ), the NSF (DMS-2054253 and IIS-2205441, HZ and JZ), Merck Biostatistics and Research Decision Sciences (HZ, JZ, and XD), and Hellman Fellowship (XD).

\section*{Disclosure Statement}
\noindent
The authors report that they have no competing interests to declare.




\bibliographystyle{apalike}
\bibliography{ref}

\end{document}